\documentclass[10pt,twocolumn,letterpaper]{article}

\usepackage{cvpr}
\usepackage{times}
\usepackage{epsfig}
\usepackage{graphicx}
\usepackage{amsmath}
\usepackage{amssymb}
\usepackage{subcaption}
\usepackage{multirow}
\usepackage{cite}
\usepackage{ctable}
\usepackage{comment}
\usepackage{tabularx} 
\usepackage{bm}
\usepackage[pagebackref=true,breaklinks=true,letterpaper=true,colorlinks,bookmarks=false]{hyperref}
\newcommand*{\rom}[1]{\expandafter\@slowromancap\romannumeral #1@}
\newcommand{\AL}[1]{{\color{black}#1}}

\cvprfinalcopy 

\def\cvprPaperID{40} 

\ifcvprfinal\fi
\begin{document}

\tolerance=999
\sloppy

\title{Debiasing Skin Lesion Datasets and Models? Not So Fast}
\author{Alceu Bissoto\textsuperscript{1} ~~~Eduardo Valle\textsuperscript{2} ~~~Sandra Avila\textsuperscript{1}\\
\textsuperscript{1}Institute of Computing (IC) ~~~\textsuperscript{2}School of Electrical and Computing Engineering (FEEC)\\ 
RECOD Lab., University of Campinas (UNICAMP), Brazil\\
}

\maketitle
\thispagestyle{empty}
\begin{abstract}
Data-driven models are now deployed in a plethora of real-world applications — including automated diagnosis — but models learned from data risk learning biases from that same data. When models learn spurious correlations not found in real-world situations, their deployment for critical tasks, such as medical decisions, can be catastrophic. In this work we address this issue for skin-lesion classification models, with two objectives: finding out what are the spurious correlations exploited by biased networks, and debiasing the models by removing such spurious correlations from them. We perform a systematic integrated analysis of $7$ visual artifacts (which are possible sources of biases exploitable by networks), employ a state-of-the-art technique to prevent the models from learning spurious correlations, and propose datasets to test models for the presence of bias. We find out that, despite interesting results that point to promising future research, current debiasing methods are not ready to solve the bias issue for skin-lesion models.

\end{abstract}

\section{Introduction}\label{sec:intro}

Models learned from data risk learning biases from that same data. When models learn the correct features, they are robust, generalizing for uncontrolled situations in the real world. Biases in the training set destroy that robustness, because models learn spurious correlations that will not be found (at least not reliably) in real-world situations. The deployment of such models for critical tasks, such as medical decisions, can be catastrophic.

As data-driven models get better and are deployed in a plethora of real-world applications — including automated diagnosis — we must understand that issue in order to trust them with those critical decisions.

For critical contexts, such as medical applications, the involved agents must be able to explain their decision process. Machine learning models deployed in those scenarios respond to those same rules — including, in several jurisdictions, legally. However — despite the surge in methods for model interpretability — we are far from explaining the predictions made by them.

Bissoto et al. \cite{bissoto2019constructing} investigated bias for skin-lesion datasets and found troubling signs, showing shockingly high performances for deep neural networks trained with images where the lesions appear occluded by large black bounding boxes. The performances were comparable to those of networks trained with \textit{additional} dermoscopic attributes. The networks were unable to exploit clinically-meaningful information in the form of dermoscopic features, neglecting those in their decision process.

Those results motivated this work, whose objective is twofold: on the one hand, we attempt to finding out what are the extraneous, spurious correlations exploited by biased networks, on the other hand, we attempt to apply techniques to \textit{debias} the models, removing such spurious correlations from them. 

To attain our first objective we analyze of the possible influence of seven artifacts present in skin-lesion images: dark corners (vignetting), hair, gel borders, gel bubbles, rulers, ink markings/staining, and patches. 
Previous works have noticed the presence of those artifacts in skin-lesion datasets and their possible effect on classification performance~\cite{bisla2019towards, winkler2019association, mishra2016overview}.  The novelty of this work, is an integrated and systematic study of those effects, in which we employed manual annotation of the presence of the artifacts to confirm our hypotheses. We annotate by hand two of the most employed skin lesion datasets: ISIC 2018 Task 1 \& 2 \cite{isic2018}, and the Interactive Atlas of Dermoscopy \cite{argenziano2002dermoscopy}. As far as we know, this level of analysis has not been attempted before. 

To attain our second objective, we employ a state-of-the-art technique to prevent the models from learning spurious correlations. Correctly assessing the performance of such techniques is not obvious: they usually lead to \textit{lower} accuracies, since the resulting models will not have the ``unfair advantage'' of the spurious correlation (advantage, of~course, that is illusory, and that would disappear if the models were to be used in actual clinical practice). One of our contributions is proposing protocols for this delicate assessment.
\AL{We have made the annotations created, the employed sets of data, and source code available\footnote{\url{https://github.com/alceubissoto/debiasing-skin}}}.

Summarizing, the main contributions in this work are:\vspace{-0.05cm}
\begin{itemize}
    \item We annotate two of the most popular datasets for skin lesion analysis with the presence of 7 visual artifacts that can lead to dataset biases.\vspace{-0.1cm}
    \item We evaluate how those artifacts affect classification models in different experiments that focus the attention of the network in different aspects of the image.\vspace{-0.1cm}
    \item We assess the capability of a state-of-the-art solution for bias removal in the skin lesion context, proposing protocols to evaluate the models in that scenario.\vspace{-0.1cm}
    \item We discuss the importance of studying bias in skin lesion analysis, and provide directions for future works to solve this key problem with our data.
\end{itemize}

\section{Related Work}
\textbf{Bias Detection:}
Bias in classification attracted much attention from researchers, who addressed the issue for datasets of different sizes and specialization levels. 

ImageNet~\cite{deng2009imagenet}, arguably the most studied large-scale dataset for image classification, present many biases despite containing more than $1$ million samples and $1,000$ classes. In ObjectNet \cite{barbu2019objectnet}, the authors provide a test-only dataset with $50,000$ images in $313$ ImageNet classes, captured by MTurkers in their own house, following special guidelines to reduce biases, in particular, randomizing backgrounds, rotations, and image viewpoints. 
Many state-of-the-art architectures for ImageNet presented a drop of performance of $\sim40$ p.p. on ObjectNet, showcasing the high impact of dataset bias.

Bias is also a concern for the task of Visual Question Answering (VQA), where models answer a textual question based on the appearance of an image. The model must make sense of the visual features to correctly answer the question, but,in some cases, may (undesirably) exploit biases in the data that enable them to answer questions without even considering the visual information. The classic example is answering ``yellow'' to ``Which color is the banana?'' in a dataset where almost all bananas are yellow. Goyal et al. \cite{goyal2017making} advanced the state of the art with purposefully debiases, and much more difficult dataset they called VQAv2 \cite{antol2015vqa}, balancing the number of answers for every question.  

Those examples of purposeful dataset debiasing provide interesting precedents, that could be explored by medical works, enabling the assessment of the actual generalization capability of solutions. However, since in medical images — and skin lesion analysis in particular — acquiring annotated data is much more expensive and laborious, literature resorts to other methods.

Bissoto et al. \cite{bissoto2019constructing} presented concerning results for skin lesion analysis, showing that even when deep neural networks are blinded from most of the relevant information, by placing a black bounding box on top of the lesion hiding at least 70\% of the image, they still make predictions much above random chance, in fact, even surpassing human specialists benchmarks in the task. 
Winkler et al. \cite{winkler2019association} also evaluated how bias affects performance, showing that ink markings/staining influences the ability of models of recognizing melanomas.

\textbf{Bias Removal:} If datasets cannot be unbiased a priori, removing biases from models may help generalization. 

Cadene et al. \cite{cadene2019rubi} proposed a solution to prevent VQA models from making decisions based only on the text query, without looking at the target image (as explained above). They add a classification head connected to the query text encoder that answers the question without using the image, and whose loss influences \textit{negatively} the overral model loss. That way, the model is encouraged to look at the samples as a whole. 

Assessing biases for VQA was largely made possible due to the existence of the VQAv2 dataset \cite{goyal2017making}, which was especially designed to have balanced answers for each question. For most applications, however, such datasets do not exist, making assessment delicate. Biased models will have better performance: the spurious correlations give them an advantage, after all. But simply striving for models with lower~performance leads, of course, to trivial uninteresting solutions. 

Often, the proposed solution is creating \textit{artificially biased} sets \cite{alvi2018turning, kim2019learning, shetty2019not}, with the training and test sets biased in opposite directions. This creates what we call here a \textbf{trap set}, since biased models, by learning spurious correlations, ``fall into the trap'', and have poor performances on test. 

So far, the most successful strategy for bias removal for classification tasks is employing auxiliary classification heads that are trained to detect features causing bias, and use them to influence the training of the main classification task. That strategy, however, requires labeling the biasing features.
Alvi et al. \cite{alvi2018turning} used domain confusion loss to unlearn the biases, i.e., for each known domain that introduce a bias (in their case, gender or age), there is a classifier that is trained to detect the bias, and a domain confusion loss that updates the feature representation to make the bias classifier perform worse.
The state-of-the-art method Learn Not To Learn (LNTL)~\cite{kim2019learning} also works with a set of known, annotated biases (such as color, gender, and age). The authors propose to use a feature extractor that feeds two classifiers: one for the main problem, the other for bias. The main classifier and the feature extractor are pre-trained to solve the target problem, and then go through an unlearning phase where the bias classifier learns the biases and backpropagates \textit{reversed} gradients to the main classifier's feature extractor, such that it unlearns the same biases. Another solution, following the same scheme, was proposed for debiasing action classification models in video \cite{choi2019can}.

\section{Data Customization} \label{sec:data}

In this section, we discuss the data employed in this work. Deep learning models are very greedy for data, and able to thoroughly exploit the available training samples. Contrarily to \AL{classic} machine-learning models, deep models learn seamlessly feature extraction and decision layers. The risk, on small datasets (such as the ones in medical tasks) is the models learning not only all relevant information, but also all spurious features present in the data, compromising their generalization to actual real-world situations.

We start by presenting the datasets \AL{commonly} used in skin lesion analysis. Next, in Sec.~\ref{sec:norm}, we propose a modified dataset to reduce the influence of bias. In Sec.~\ref{sec:annotation}, we discuss our manual annotation of both the ISIC 2018 Task 1 \& 2 \cite{isic2018} and the Interactive Atlas of Dermoscopy \cite{argenziano2002dermoscopy} for artifacts that could lead to spurious correlations on the data, and provide an analysis of the correlations we found among those artifacts and the target labels. Finally, in Sec.~\ref{sec:trap}, we describe our \textit{trap sets}, i.e., datasets whose training and test splits present high and  opposite correlations between the annotated artifacts and target label. Armed with those data, we assess the bias removal procedures described in Sec.~\ref{sec:biasremoval}.

\subsection{Traditional datasets} \label{sec:traddata}

Research of skin lesion analysis relies heavily on two public datasets: the ISIC Archive \cite{isicarchive} and the Interactive Atlas of Dermoscopy \cite{argenziano2002dermoscopy}. 

The ISIC Archive, associated with the ISIC Project and the ISIC Challenge \cite{isic2016, isic2017, isic2018}, is an ongoing cooperation, with a growing number of samples, diversity of classes, and metadata annotation. 
In this work, we used the 
ISIC 2018 Task 1 \&~2 subset of the Archive, used in the lesion segmentation (task 1) and dermoscopic attribute segmentation (task 2) of the 2018 ISIC Challenge. That subset is helpful for our bias study since every image has ground-truth segmentation masks for the lesion and dermoscopic attributes. The dataset is composed of $2,594$ dermoscopic images of 3~different classes: melanoma (malignant), nevus, and seborrheic keratosis (both the latter benign).

The Interactive Atlas of Dermoscopy \cite{argenziano2002dermoscopy} (Atlas, for short) is an educational resource used to train dermatologists. Because of the pedagogical purpose, the dataset has several difficult-to-diagnose cases, exceptions to the typical trends, hard to identify for both humans and machines. 
A drawback of Atlas is the lack of lesion segmentation masks, which we solve by using segmentation masks generated by Bissoto et al. \cite{bissoto2019constructing} using SegAN \cite{xue2018segan}. On the other hand, the Atlas provides both dermoscopic and clinical images for the case, and the latter allow a very challenging assessment of the generalization abilities of models learned on dermoscopic images.
The Atlas has $872$ dermoscopic images in the classes melanoma, nevus, and seborrheic keratosis, a subset we call here Dermoscopic Atlas. 
In those same three classes, the Atlas has $839$ clinical images, a subset we called Clinical Atlas. All cases present in Clinical Atlas are present in Dermoscopic Atlas.

\subsection{Normalized-background dataset} \label{sec:norm}

In this work, we consider as foreground in the lesion images the area occupied by the lesion itself (as delimited by a ground-truth segmentation mask, or inferred by a segmentation model), and as background everything else, which may include zones of normal skin, hair, physical artifacts like patches, and image artifacts like vignettes and shadows. The background may influence the results of skin-lesion classification, often in the form of undue learning of spurious correlations. 

In order to evaluate such effects, we propose modifying the dataset by erasing the background. In order to minimize the perturbations to the model, instead of simply using a constant value, we first learn the average image in the training set, and replace each pixel in the background by the corresponding pixel in that \textit{pixel-average training set image}. The foreground pixels are left untouched.

\begin{figure}[t]
    \centering
    \begin{subfigure}[b]{0.23\linewidth}
    \includegraphics[width=\linewidth]{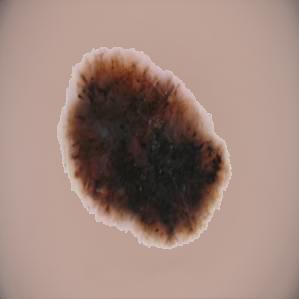}
    \caption{\AL{Traditional}}
    \end{subfigure}
    \begin{subfigure}[b]{0.23\linewidth}
    \includegraphics[width=\linewidth]{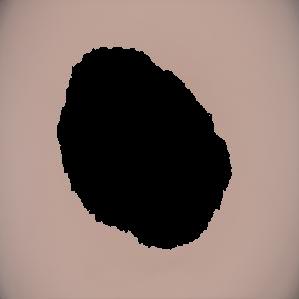}
    \caption{\AL{Skin Only}}
    \end{subfigure}
    \begin{subfigure}[b]{0.23\linewidth}
    \includegraphics[width=\linewidth]{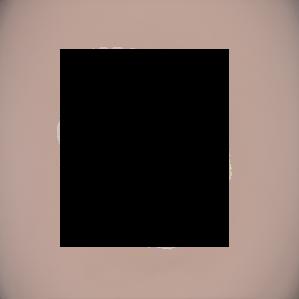}
    \caption{Bbox}
    \end{subfigure}
    \begin{subfigure}[b]{0.23\linewidth}
    \includegraphics[width=\linewidth]{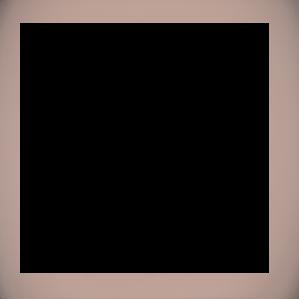}
    \caption{Bbox70}
    \end{subfigure}
    \caption{Normalized-background images. The aim of this dataset is to reduce the influence of the background, allowing to measure how misleading visual patterns in the background influence the classification.\vspace{-0.25cm}}
    \label{fig:normalizedimages}
\end{figure}

In test time, the same procedure is applied to the images, always using the average computed on the \textit{training} set. That way, images from both train and test have the same non-informative background, but only images from the training set influenced the normalized background (see Fig.~\ref{fig:normalizedimages}).

To measure the effect of this change in deep learning models, we used an Inceptionv4, following the same hyperparameters, training procedures, and data splits as \cite{bissoto2019constructing}, to allow the comparison between their results and ours. Those results appear in Table \ref{tab:normbackground}. The image disturbances (\emph{Traditional}, \AL{\emph{Skin Only}}, \emph{Bbox}, \emph{Bbox70}) are also the same as theirs.

\begin{table*}[t]
\small
\centering
\begin{tabular}{l c c c c}
      Dataset  & Traditional (\%) & Skin Only (\%) & Bbox (\%) & Bbox70 (\%) \\ \hline
    ISIC \cite{bissoto2019constructing} & $86.3 \pm 1.6$ & $77.3 \pm 1.6$ & $77.1 \pm 1.8$ & $71.1 \pm 1.6$ \\
    ISIC Normalized & $81.5 \pm 1.2$ & $72.7 \pm 1.6$ & $67.0 \pm 2.5$ & $59.8 \pm 2.1 $ \\
    \hline
    Cross-dataset \cite{bissoto2019constructing} & $83.5 \pm 0.9$ & $72.3 \pm 1.1$ & $71.3 \pm 1.8$ & $71.5 \pm 0.7$ \\
    Cross-dataset Normalized & $77.1 \pm 1.3$ & $69.0 \pm 1.1$ & $67.2 \pm 3.9$ & $64.1 \pm 0.3 $ \\
    \end{tabular}
    \caption{Results (in AUC, Area Under the ROC Curve) for the experiments with the normalized background. All the images of train and test have the same background. We run each experiment 10 times, using the same sets of images from \cite{bissoto2019constructing} to make the comparison fair.
    The \emph{Bbox70} and \emph{Bbox} sets are the most affected, since the large foreground occlusions make those models very dependent on the background. 
    }
    \label{tab:normbackground}
\end{table*}

When train and test are splits of the same dataset (ISIC), \emph{Bbox70} and \emph{Bbox} suffer considerably more ($11.3\%$ and $10.1\%$, respectively) than \emph{Traditional} or \emph{Skin Only} ($4.8\%$ and $4.6\%$), indicating that for models learned with disturbed images the background becomes much more important. The network expects background features to be visible to help it make sense of the label. 
These results showcase that the models will take advantage of all and any available information, including spurious correlations that may harm the model's ability to generalize. To avoid that, we can (and should) control the information fed to the network. 

The cross-dataset design (training on ISIC and testing on Dermoscopic Atlas) shows many of the same general trends, with  two main differences. First, the performance, even for \emph{Traditional} images with preserved background was already worse, showcasing the difficulty of the model to generalize to this different, and challenging dataset. Second, the performance drops on the normalized background case were more spread between the progressive steps, with \emph{Traditional} images already facing a heavy drop. The smoother drop on the other steps may come from the imperfect, inferred segmentation masks, instead of the ground-truth ones used on ISIC.

\subsection{Artifacts annotation} \label{sec:annotation}

For deployment in the real-world, medical tasks like patient screening or triage require not only accurate, but also reliable models, robust to variability. However, the data available to develop those models are often limited and class-unbalanced, fostering the desire to exploit every sample as much as possible. Throwing away the background — even if it results in less biases — goes against such desire: there might still be cogent, legitimate information in the skin around the lesion, providing actual context for the diagnosis. If possible, we would like to have ways to isolate bias without throwing away so much information.

\begin{figure}[t]
    \centering
    \begin{subfigure}[b]{0.3\linewidth}%
    \includegraphics[width=\linewidth, height=0.70\linewidth]{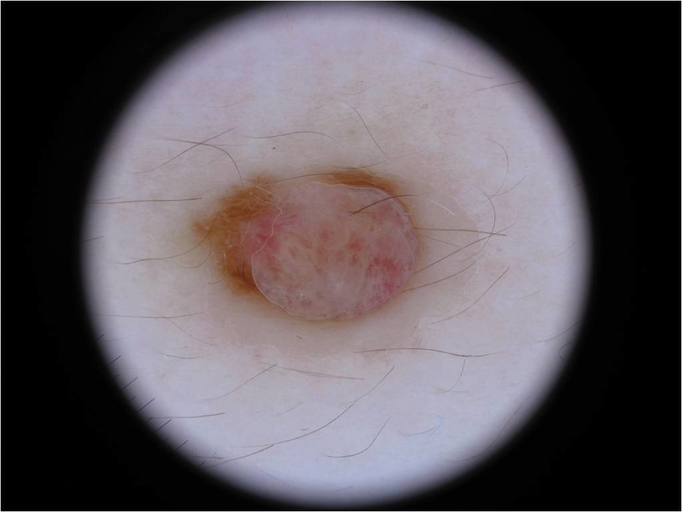}
    \caption{Dark Corners}
    \end{subfigure}
    \begin{subfigure}[b]{0.3\linewidth}%
    \includegraphics[width=\linewidth, height=0.70\linewidth]{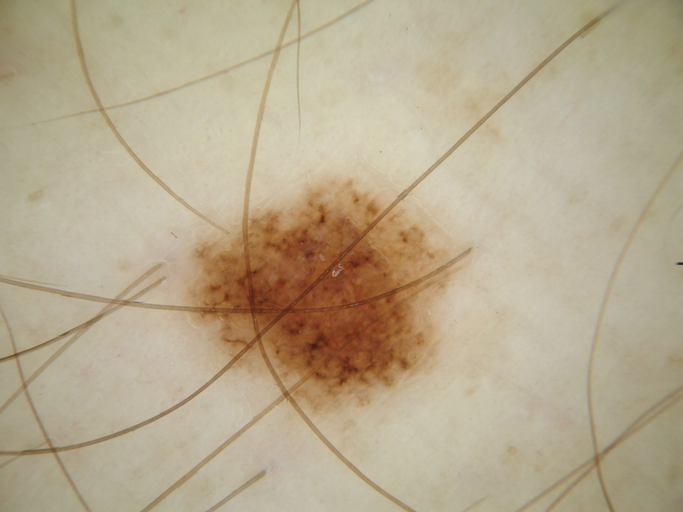}
    \caption{Hair}
    \end{subfigure}
    \begin{subfigure}[b]{0.3\linewidth}%
    \includegraphics[width=\linewidth, height=0.70\linewidth]{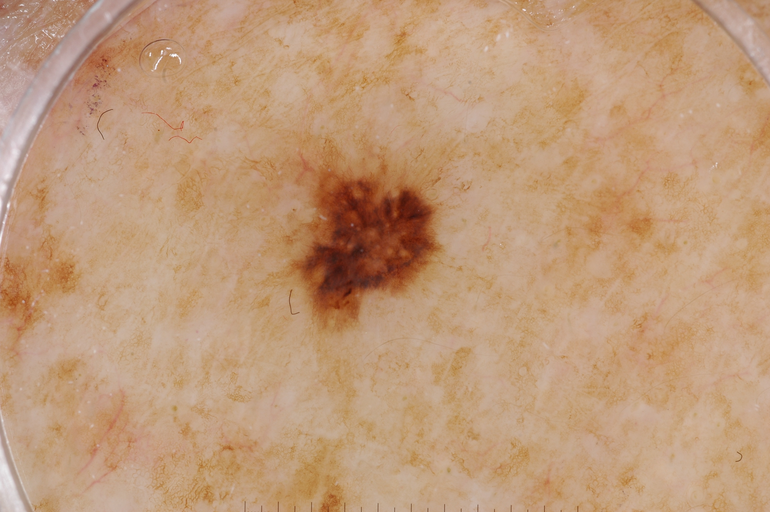}
    \caption{Gel Border}
    \end{subfigure}
    \begin{subfigure}[b]{0.3\linewidth}%
    \includegraphics[width=\linewidth, height=0.70\linewidth]{cvpr2019AuthorKit/latex/images/data/ISIC_0013775.png}
    \caption{Ruler\\ ~}
    \end{subfigure}
    \begin{subfigure}[b]{0.3\linewidth}%
    \includegraphics[width=\linewidth, height=0.70\linewidth]{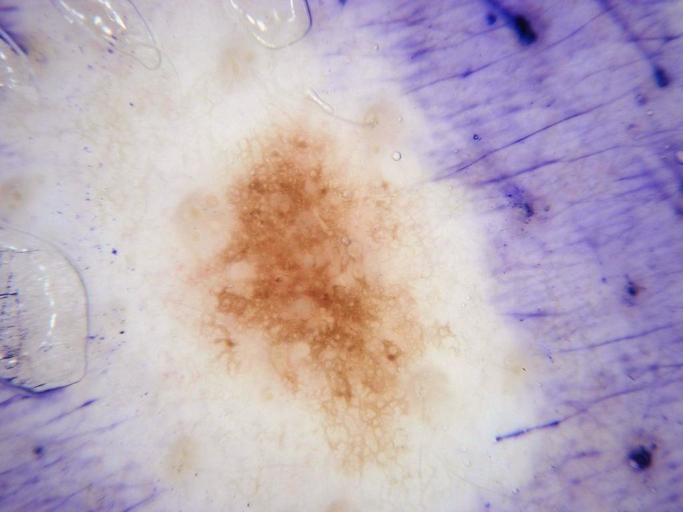}
    \caption{Ink markings and Gel bubbles}%
    \end{subfigure}
    \begin{subfigure}[b]{0.3\linewidth}%
    \includegraphics[width=\linewidth, height=0.70\linewidth]{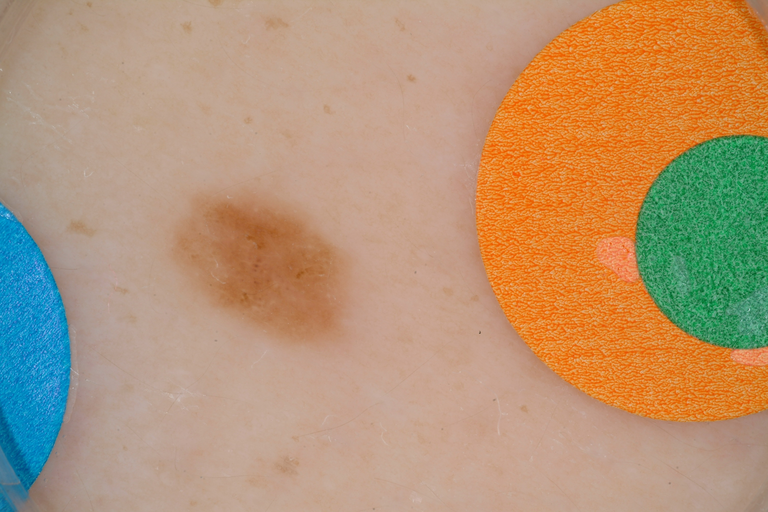}
    \caption{Patches \\ ~}
    \end{subfigure}
    \caption{The 7 possible ``culprit'' artifacts selected for annotation in our datasets.}
    \label{fig:artifacts}
\end{figure}

To attempt just that, we selected $7$ possible ``culprit'' artifact for creating bias: dark corners (vignetting), hair, gel borders, gel bubbles, rulers, ink markings/staining, and patches applied to the patient skin (Fig.~\ref{fig:artifacts}). We manually annotated the $2,594$ images of ISIC 2018 Tasks 1 \& 2, and $872$ images of Dermoscopic Atlas. 

\begin{figure}
    \centering
    \includegraphics[trim=1cm 5.25cm 1cm 5.05cm, clip, width=0.87\linewidth]{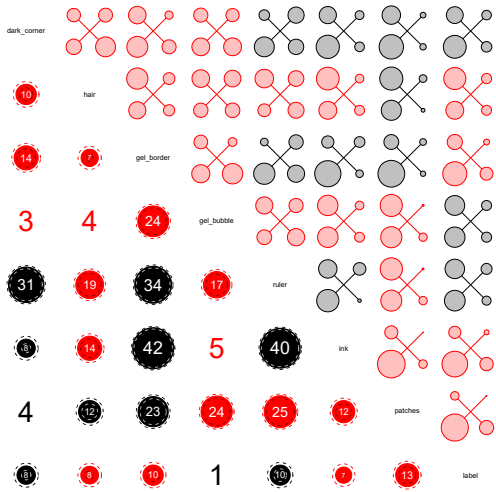}
    \caption{    Correlogram of annotated artifacts and target labels (benign and malign). The variables are shown on the diagonal. Spearman correlations shown in lower triangle, and actual data distribution shown in upper triangle (black for positive, and red for negative correlations).}
    \label{fig:correlogram_data}
\end{figure}

However, when we attempt to correlate those annotated artifacts to the target labels (malignant and benign), we find that the correlations are modest. Fig.~\ref{fig:correlogram_data} shows the correlation analysis, with the variables on the diagonal and Spearman correlations ($\rho$) in the lower triangle (black for positive, and red for negative correlations). The filled circles' areas are proportional to $\rho$, and the dashed circles's to the 95\%-CI. If the CI contains zero, we omitted the circles. The pictograms in the upper circle show the joint distribution of the two variables, with the area of each small circle in the cross proportional to the amount of samples in the dataset in a given combination.
The lack of strong individual correlations suggests the possibility the models are able to extract and combine weak correlations from several sources to arrive at a combined considerable bias — showcasing the danger of cumulative small bias. Other possibility, which we find more probable, is that bias in the data is an insidious phenomenon, and that the actual ``culprits'' may be difficult to find. Subtle differences in acquisition equipment or procedure, for example, may appear impossible for humans to detect, but very easy for machines to exploit. Notice that those two possibility are \textit{not} mutually exclusive.

But can the network make sense of this information, if we encourage it to? We have seen before, during the ISIC 2018 Challenge Task 2, that it can be very hard for networks to verify the presence of dermoscopic attributes, for example, making semantic segmentation results for this task very low (the best solution achieved~$\approx 0.3$ in the Jaccard metric\AL{\footnote{\url{https://challenge2018.isic-archive.com/leaderboards}}}
). However, differently from the annotated artifacts, dermoscopic artifacts can provide useful and correct correlations to the network.

To measure the ability of neural network to learn those artifacts, we built binary classifiers for each one of them, using the same model architecture and hyperparameters explained in Sec.~\ref{sec:norm}. The results appear in Table \ref{tab:artifacts_classification}, where the high performance for all the artifacts highlight a concerning ability of the network to correctly identify the artifacts even on highly disturbed sets such as \emph{Bbox90}, where $90\%$ of the pixels of the image are occluded. 

\begin{table*}[t]
\small
    \centering
\begin{tabular}{l  c  c  c  c  c  c  c}
      ~  & Dark Corner (\%) & Hair (\%) & Gel Border (\%) & Gel Bubble (\%) & Ruler (\%) & Ink (\%) & Patches (\%) \\ \hline
    \multicolumn{8}{c}{ISIC} \\ \hline
    Traditional & $95.6$ & $94.0$ & $93.4$ & $85.3$ & $98.2$ & $97.8$ & $98.2$ \\
    Bbox90      & $80.7$ & $79.3$ & $79.6$ & $71.3$ & $88.1$ & $81.4$ & $97.9$ \\ \hline
    \multicolumn{8}{c}{Cross-dataset ISIC/Atlas} \\ \hline
    Traditional & $86.0$ & $87.0$ & - & $79.5$ & $70.3$ & $75.7$ & - \\
    Bbox90      & $92.7$ & $71.4$ & - & $63.3$ & $62.3$ & $50.1$ & - \\ 
    \end{tabular}
    \caption{Results (in AUC) for separate classifiers trained to detect each of the 7 annotated artifacts. The classifiers are able to achieve high performances even in the very disturbed image set \emph{Bbox90}, and on the cross-dataset scenario.}
    \label{tab:artifacts_classification}
\end{table*}

Also, there is a noticeable difference between the artifacts distributions of ISIC and Atlas datasets. Not only~Atlas does not present Gel borders nor Patches, but even though the performances are still highly predictive, in the cross-dataset experiment they are significantly lower. This shows how important it is to report cross-dataset results, and to make publicly available different and more diverse datasets to test the robustness and quality of the solutions.

\subsection{Trap sets} \label{sec:trap}

In this work, we have two objectives: 1) we want to understand the sources of bias, try to quantify and unravel their connection to features and input data; 2) we want to remove the identified bias from skin lesion datasets.
The removal of bias is a challenging problem partly because we need to control features that are entangled inside our model, but also because it is hard to quantitatively show the benefits of the removal of this bias. It would not be unusual if an unbiased network performed worse than a biased one in the same test set. Of course, if a set of spurious correlations were exploited by the network, it is because they were helping the loss to decrease, and to reach higher values in the metrics. The model takes advantage of spurious correlations because of its beneficial int the minimization problem point of view. The problem is that it also harms generalization, which is a lot harder to measure.

To try to tackle this problem, bias removal solutions usually resort to manipulated datasets where the bias is amplified enough so the networks are forced to exploit handcrafted high correlations between the target labels, and the bias \cite{kim2019learning}. We followed this procedure and created a training and test set where the correlations are amplified between our artifacts and the malignancy of the skin lesion, at the same time that correlations in train and test are \textbf{opposite}.

\section{Bias Detection and Removal} \label{sec:biasremoval}

In critical contexts, such as the medical one, it is very important to understand the decision path of its agents. Doctors' decisions are based on several studies, and experiences of previous cases. For AI, it only has the experience of the previous cases it has seen. We do not control what aspects are more important, or provide a guideline for the algorithm to follow to achieve the best performance. The patterns found can be the same as used by human experts, or they can be something completely different that only makes sense in the high dimensions where the data are transformed over and over.

The shocking results by Bissoto et al. \cite{bissoto2019constructing}
showed convolutional neural networks achieving expert-level performance on skin lesions analysis without actually feeding the lesion to the model. The authors placed a bounding box on top of the lesion, covering at least 70\% of the image, and still the performance maintained. We extend the investigation of bias, trying to have a better understanding of the classification models, while also investigating ways to create more robust classifiers.

\subsection{What features are being used by our models?}

\begin{figure*}[t]
    \centering
    \begin{subfigure}[b]{0.25\linewidth}
    \includegraphics[width=\linewidth]{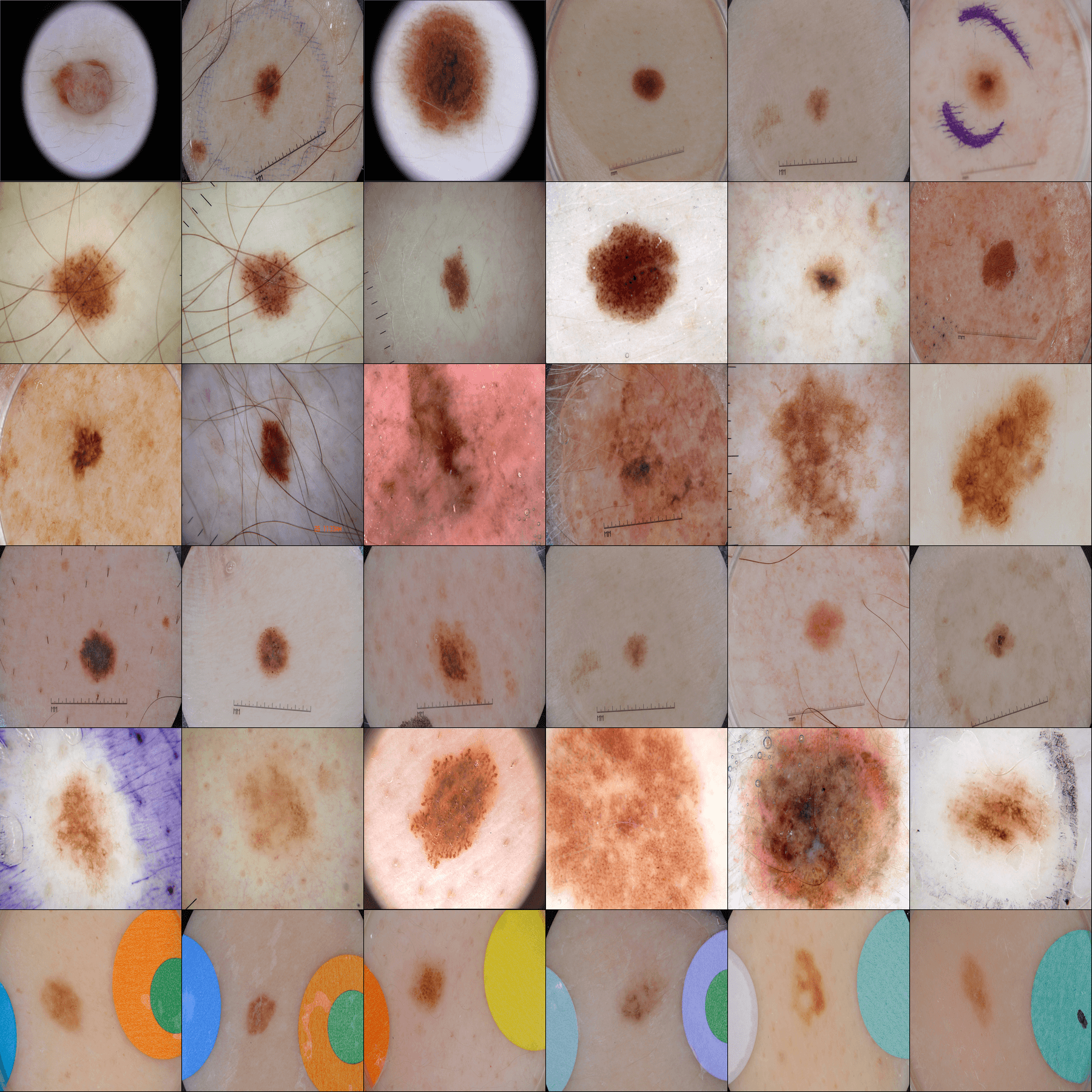} \caption{Traditional}
    \end{subfigure}
    \begin{subfigure}[b]{0.25\linewidth}
    \includegraphics[width=\linewidth]{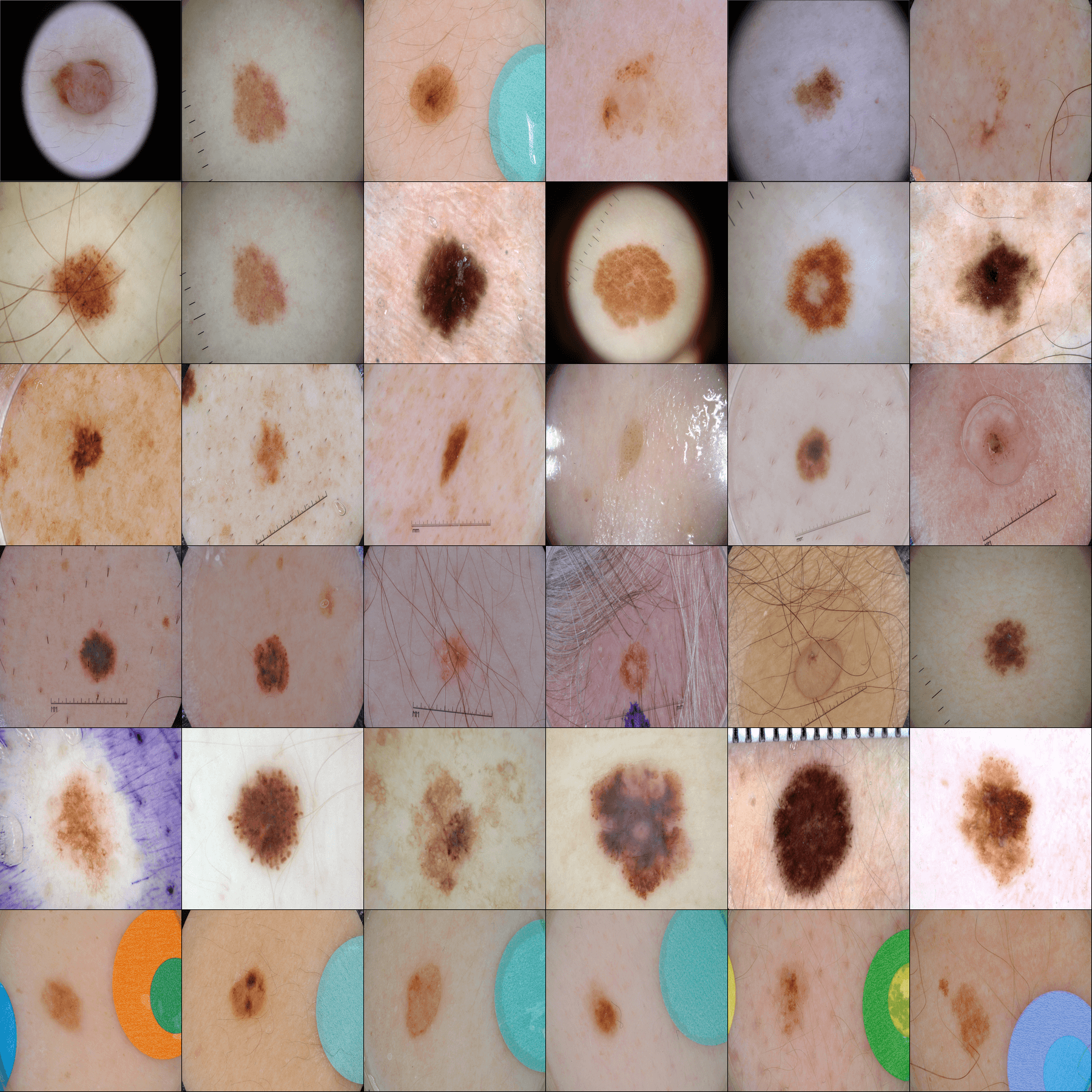}
    \caption{Normalized Bbox}
    \end{subfigure}
    \begin{subfigure}[b]{0.25\linewidth}
    \includegraphics[width=\linewidth]{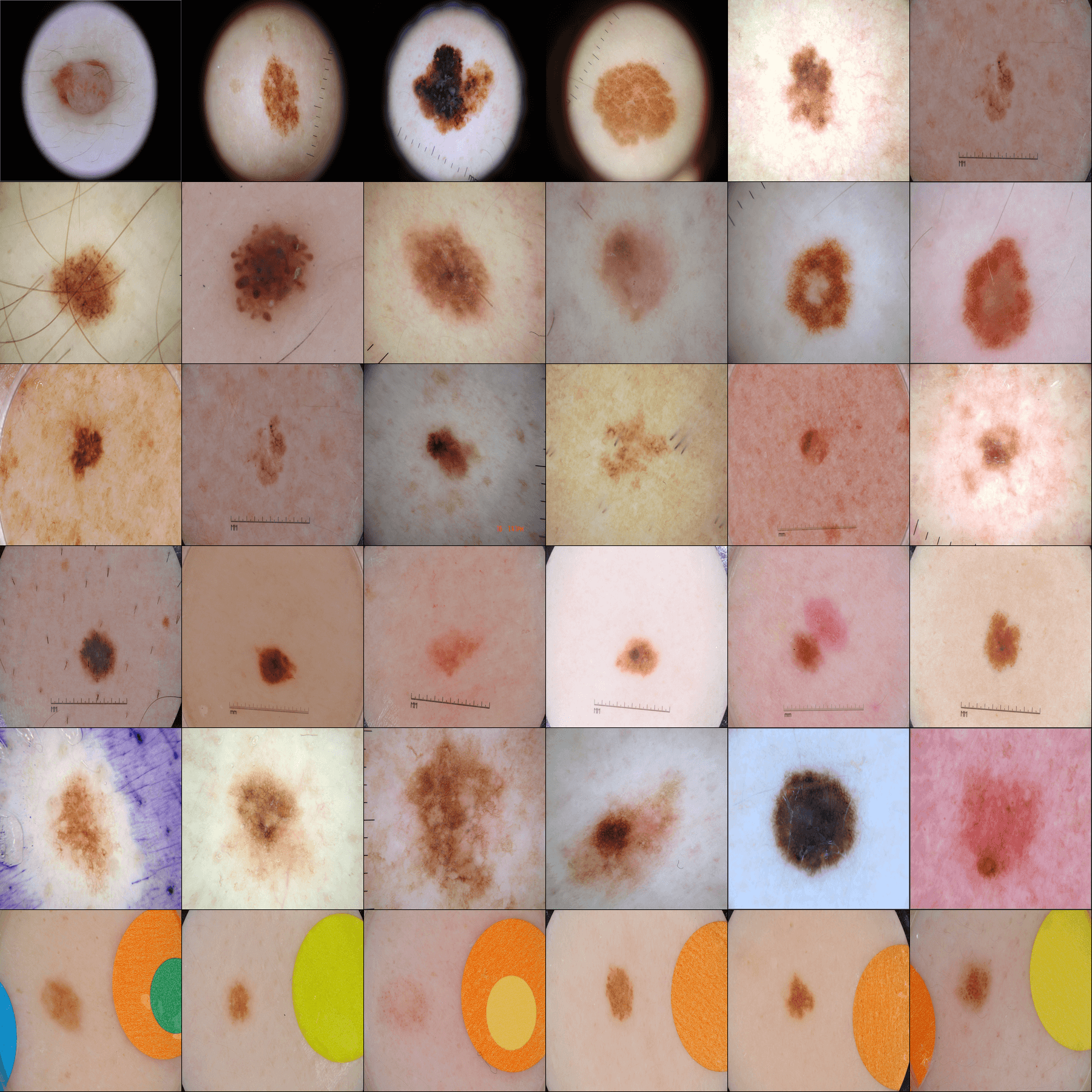}
    \caption{Bbox}
    \end{subfigure}
    \caption{Grid showing image similarity according to the features extracted by our classification model. The first column of each grid is the query, and the remaining columns are ranked according to euclidean distance of the images' features. We selected queries carefully to show different artifacts. In sequence, dark corners, hair, gel border, ruler, ink markings and patches. For clarity, we show the original images for all the cases, but the networks were trained and evaluated using their respective image type (Traditional, Normalized Bbox, and Bbox).}
    \label{fig:grids}
\end{figure*}

The first idea to understand the network, and make sense of the features being extracted and used to make the predictions is to use a visualization technique. However, the methods available in the literature failed because the architecture was complex (guided backpropagation \cite{springenberg2014striving}), or because the resultant saliency maps were too coarse (GRADCAM \cite{selvaraju2017grad}). Also, occlusion methods such as \cite{fong2017interpretable} may not be appropriate for binary problems such as ours (benign or malignant), where Gaussian perturbations (blur) can have low to no effect in the network's performance, and more aggressive perturbations (such as changing the pixel value for a solid color) can introduce some uncertainty or unknown patterns that will confuse the network and will not highlight the really important parts.

To overcome those challenges, we use a qualitative method to start understanding what information can be important to the network. We extracted features for all the images in the dataset, and ranked them with respect to Euclidean distance to a carefully selected query. The selected queries present some suspect features that the network should not take advantage of, such as dark corners, hair, ink markings and patches (see Fig.~\ref{fig:artifacts}). By comparing the retrieved lowest distance images, we can identify some of the features the network is learning (see Fig.~\ref{fig:grids}).

By analyzing the grids, we see that \AL{the network trained with traditional images} is able to explore features that are not directly related to artifacts. For example, instead of learning the presence or absence of dark corners ($1$st row), and ink markings ($5$th row), it learns to detect oval silhouettes that can be made by dark corners, gel bubbles, ink markings or even bigger lesions' borders. Looking at the overall features of the lesions shows that the network rank images similarity based on diversified sources of correlations, where each line usually contains lesions of different sizes, placed in different positions in the image, and presenting different artifacts.

The next two grids contain the ranking of images of networks trained with black bounding boxes on top of the lesions. For clarity, we display the original lesions in the grids. Both grids share some similarities that are bound to the bounding box training: the importance of lesion size and positioning have increased drastically.
In contrary to the \AL{training with traditional images}, there are no large lesions among the small ones, and vice-versa. 
The positioning is also important, especially for the normalized case. A clear example is that even without seeing the patches, the network grouped together images that contain them because they are in the left portion of the image. The normalized images also present diversified artifacts in most rows. Differently from both \emph{Traditional} and \emph{Bbox}, we can see dark corners and patches in the middle of the grid.
The artifacts are also much more important for the \emph{Bbox} case, where dark corners containing lesions of similar size are ranked first than other cases. Also, the network identifies hair correctly, grouping them, and the patches selected contain similar colors.

\subsection{Bias removal}

We want our models not to learn about the selected artifacts, once they can harm its generalization ability.

We employ Kim et al. \cite{kim2019learning} state-of-the-art method called Learning Not To learn (LNTL) for removing bias, and apply to our skin lesion analysis problems. 
The proposed framework is composed of three main components:\vspace{-0.13cm}
\begin{itemize}
    \item A feature extractor, responsible for providing useful features to all classification heads.\vspace{-0.13cm}
    \item The main task classification head, responsible to make sense of the extracted features to solve the target problem. In our case, this problem is skin lesion classification into benign or malignant.\vspace{-0.13cm}
    \item The bias classification heads. Each classification head is focused on a component responsible for injecting spurious correlations in the data. In our case, we have one classification head for each of the annotated artifacts: dark corners, hair, gel borders, gel bubbles, rulers, ink markings, and patches.
\end{itemize}

The training of the solution has two phases:
First, there is a pretraining phase where only the feature extractor and the main task classification head are trained. Both are optimized to solve the task of classifying skin lesions into benign or malignant.
Next, we load the best weights for~both according to the validation loss, and the bias classification heads come into action. Now, the feature extractor, main task classifier, and bias classification heads are trained. The feedback received by the feature extractor from the bias classification heads is reversed (negated). Thus, the feature extractor is getting worse at extracting bias information, while the classification heads are getting better at detecting the presence of the artifacts in the extracted features. 

To highlight the method of bias removal, we use trap datasets (see Sec.~\ref{sec:trap}) where the correlations between the artifacts (bias) and the labels are amplified, and also are opposite between training and test splits.

\subsection{Implementation details}

We follow the guidelines from Kim et al. \cite{kim2019learning}, and use a ResNet18 network \cite{he2016deep} as our main network. Because it is not as deep as other architectures used for skin lesion analysis \cite{bissoto2018skin, valle2018data, solo_or_ensemble}, we also experiment with a ResNet152 architecture, which is capable of achieving performance close to the state-of-the-art. By sticking with ResNet architectures, we can follow the same procedures proposed by Kim et al., eliminating the uncertainty that comes when adapting the original solution for a new family of architectures. 

The first two major blocks from the ResNet architecture are used as feature extractor. The classification head for the main task of discriminating skin lesions into benign and malignant, is composed of the reminiscent two blocks and a Dense layer. 
Each classification head responsible for predicting the presence of the annotated artifacts is composed of a single linear layer. 
All classification heads, both the artifacts' and the main task's heads, are fed with the same extracted features. To unlearn features on the extractor, we inverse the gradients and multiply it by a factor of $0.3$. We use SGD with a momentum of $0.9$ and weight decay as $0.0005$. 
For all networks, we decrease the learning rate by a factor of $10$ after $40$ epochs, and use the cross entropy loss.

The pretraining phase lasts for $100$ epochs, and the best validation model is selected to go through the bias removal phase, which also lasts for 100 epochs. We report the result achieved by the end of the $100$th epoch. We use data augmentation for all training phases: we apply random horizontal and vertical flips, random resized crops that contain from $75\%$ to $100\%$ of the original image, random rotations between $-45$ and $45$ degrees, and random hue changes between $-20\%$ to $20\%$.
We apply the same augmentations on both train and test. For the evaluation, we average the predictions over $50$ augmented versions of each image. We normalize the input using the computed ImageNet's training set mean and standard deviation.

\subsection{Results and discussion}

We want our networks to be less biased toward the detected artifacts.
In Table \ref{tab:results_removal} we present the performance of our bias removal experiments. 
\begin{table*}[h]
\small
    \centering
    \begin{tabular}{ccccc}
         Experiment                  & Architecture & Trap Test ($\%$)  & Atlas Dermoscopic ($\%$)  & Atlas Clinical ($\%$) \\ 
         \hline
        \AL{Unchanged}                 & Inceptionv4 & $52.6 \pm 1.8$       & $\bm{78.5 \pm 1.6}$                  & $63.4 \pm 1.1$ \\
        Normalized                  & Inceptionv4 & $\bm{55.8 \pm 2.4}$  & $72.4 \pm 1.2$                  & $-$ \\
        LNTL \cite{kim2019learning} & ResNet152   & $54.5 \pm 3.0$       & $78.4 \pm 0.8$                  & $\bm{70.1 \pm 1.1}$ \\
         \hline
        \AL{Unchanged}                 & ResNet18    & $44.7 \pm 1.5$       & $72.2 \pm 2.1$                  & $65.8 \pm 1.2$ \\
        Normalized                  & ResNet18    & $\bm{62.4 \pm 3.3}$  & $70.5 \pm 1.0$                  & $-$ \\
        LNTL \cite{kim2019learning} & ResNet18    & $51.4 \pm 1.7$       & $\bm{76.0 \pm 0.9}$                  & $\bm{68.2 \pm 2.4}$ \\
    \end{tabular}
    \caption{Result (in AUC) of the bias removal solution. We use 5 splits of data, that are kept the same through all experiments. Also, we use augmentation on test with 50 samples for more reliable predictions.}
    \label{tab:results_removal}
\end{table*}{}

First, it is noticeable the diagnosis difficulty introduced by the trap sets. With the correlations between the artifacts and diagnosis amplified, the network starts relying on this information to make predictions. Because of this, even though the models achieve very high performances on training and validation sets, the metrics collapse in the test set. 

The results also show how difficult it is for the method to diminish the influence of bias in our solution. We use LNTL with different architectures to attempt to remove bias in our solution. 
The low performance of LNTL, which is the current state-of-the-art for bias removal, shows how entangled the artifacts and the diagnostic label of the lesions can be. The highest increase in performance happened for clinical Atlas. We think that this higher improvement for clinical images is due to the difference in the distribution between clinical and dermoscopic images. For a solution trained with dermoscopic images to also present high performance with clinical images, it requires better generalization. This way, the debiasing of our models may be necessary when applying automated skin lesion analysis in the real world.

The low performance of the normalized dataset using the trap dataset is one more evidence of how hard it is to deal with bias in the pixel dimension. Artifacts like hair, gel bubbles, and even rulers can be very difficult to remove from images since these artifacts are often displayed on top of the lesions. This way, methods that modify the background are not enough to fully remove their influence. 
We think that bias needs to be dealt with in the feature space, disentangling artifacts from the diagnostic label. 

\subsection{Failure attempts}

Since we are considering a more difficult task than the one attempted by the LNTL authors, we also considered some modifications. All of the listed attempts in the following have not resulted in any improvements in our results, but we think it is valid to list them so future researchers can continue investigating this problem.\vspace{-0.15cm}  

\begin{itemize}
    \item Experiments with the larger ISIC 2019 training dataset~\cite{tschandl2018ham10000,combalia2019bcn20000}. Our intuition was that more data could enable bias classification heads and feature extractors to distill more relevant information, which would benefit the debiasing procedure. In this case, we took advantage that the individual bias classifiers described in Sec.~\ref{sec:annotation} achieved very high performance, and we inferred the present artifacts for all the $25,331$ images. Since most of those images also do not have their annotated segmentation masks, we inferred those too with a state-of-the-art segmentation network \cite{deeplabv3+} trained with images from the ISIC Archive \cite{isicarchive}.\vspace{-0.15cm}
    \item The ResNet \cite{he2016deep} architecture can be divided into four major blocks, each containing ResNet blocks that contain multiple convolution layers each. The original architecture implemented by the LNTL authors is the ResNet18, which contains two ResNet blocks inside each of the four major blocks. The deeper ResNet152 network is not ``symmetric'' as ResNet18, so we tried two configurations: after the second, and after the third major block. This way, the weight updates caused by the bias classification heads can change more complex concepts located at the middle and end of the network.\vspace{-0.15cm} 
    \item Deeper classification heads, expecting that if they are better at finding correlations at the extracted features, they can provide more meaningful gradients to the feature extractor and change more important weights.
\end{itemize}

\section{Conclusion}\label{sec:conclusion}

Our work shows how models with traditional and disturbed inputs use different features in order to learn unwanted biases. We show that state-of-the-art methods for bias removal are not ready to cope with those bases — at least when facing our challenging trap sets.

To understand the features used, we verified two different behaviors. For traditional inputs, our networks exploit complex and diversified correlations to map the pixels to the diagnostic. For disturbed sets, the network is still able to make sense of spurious correlations in the data, such as lesion position and lesion size. Despite size is a valid characteristic for dermatologists when analyzing skin lesion images, there are no guidelines followed by image acquisition that keep the sizes comparable among different cases. 

Our results showed how difficult it can be to understand the features used by the network, and to make it interpretable. Note that interpretability can be decisive for the adoption of automated skin lesions, or enable it to aid doctors in difficult cases. 
When attempting to remove bias, the state-of-the-art method was only able to achieve an improvement over the very difficult, and out of the data distribution, Atlas clinical dataset.

Future works should consider more diverse images from different sources, dermoscopic and clinical, and with different diagnosis. Diversity in train and, especially, test sets will lead to more robust and reliable solutions. 

We must be careful about the information we feed to our models, since data-driven models will exploit every correlation available to minimize their loss functions, without any concern about clinical plausibility. Further studies to interpret those black boxes, and control the information used are crucial. We believe that domain adaptation and representation learning will contribute to those developments: working with multiple and diversified data will lead us to deal with dataset shifts, while the ability to map our images to a controlled space, where the features are disentangled and known, will allow us to select unbiased domains out of biased ones, and learn how to extract unbiased features to compose world-class diagnostic systems.

\section*{Acknowledgments}
A. Bissoto is partially funded by CAPES (88887. 388163/2019-00). S. Avila is partially funded by~FAPESP (2017/16246-0, 2013/08293-7). A.~Bissoto and S.~Avila are also partially funded by Google LARA~2019. E. Valle is partially funded by a CNPq PQ-2 grant (311905/2017-0), and by a FAPESP grant (2019/05018-1). This~project is partially funded by a CNPq Universal grant (424958/2016-3). RECOD~Lab. is supported by projects from FAPESP, CNPq, and~CAPES.
We acknowledge the donation of GPUs by NVIDIA. We thank João Cardenuto for providing the annotation~tool.
{
\bibliographystyle{ieee_fullname}
\bibliography{arxiv}
}

\end{document}